\documentclass[conference]{IEEEtran}
\IEEEoverridecommandlockouts

\usepackage{url}
\usepackage{subcaption}

\usepackage{tabularx,calc}

\usepackage{colortbl}
\usepackage[table]{xcolor}

\usepackage[colorlinks]{hyperref}
\hypersetup{
    colorlinks=true,
    linkcolor=blue,
    filecolor=blue,      
    urlcolor=blue,
    citecolor=cyan,
}

\usepackage{cite}
\usepackage{amsmath,amssymb,amsfonts}
\usepackage{algorithmic}
\usepackage{graphicx}
\usepackage{textcomp}
\usepackage{xcolor}
\def\BibTeX{{\rm B\kern-.05em{\sc i\kern-.025em b}\kern-.08em
    T\kern-.1667em\lower.7ex\hbox{E}\kern-.125emX}}
\begin{document}

\title{FireRisk: A Remote Sensing Dataset for Fire Risk Assessment with Benchmarks Using Supervised and Self-supervised Learning\\
}


\author{
    \IEEEauthorblockN{Shuchang Shen\IEEEauthorrefmark{1}, Sachith Seneviratne\IEEEauthorrefmark{1}\IEEEauthorrefmark{2}, Xinye Wanyan\IEEEauthorrefmark{1}, Michael Kirley\IEEEauthorrefmark{1}}
    \IEEEauthorblockA{\IEEEauthorrefmark{1} Faculty of Engineering and Information Technology, The University of Melbourne, Parkville, Victoria, Australia}
    \IEEEauthorblockA{\IEEEauthorrefmark{2} Melbourne school of design, The University of Melbourne, Parkville, Victoria, Australia}
    \IEEEauthorblockA{\{chuchangs\}@student.unimelb.edu.au, \{sachith.seneviratne, x.wanyan, mkirley\}@unimelb.edu.au}
}

\maketitle

\begin{abstract}
    In recent decades, wildfires have caused tremendous property losses, fatalities, and extensive damage to forest ecosystems.
    Inspired by the abundance of publicly available remote sensing projects and the burgeoning development of deep learning in computer vision, our research focuses on assessing fire risk using remote sensing imagery.
    
    In this work, we propose a novel remote sensing dataset, FireRisk, consisting of 7 fire risk classes with a total of 91,872 labelled images for fire risk assessment.
    This dataset is labelled with the fire risk classes supplied by the Wildfire Hazard Potential (WHP) raster dataset~\cite{dillon2020wildfire}, and remote sensing images are collected using the National Agriculture Imagery Program (NAIP)~\cite{maxwell2017land}, a high-resolution remote sensing imagery program.
    On FireRisk, we present benchmark performance for supervised and self-supervised representations, with Masked Autoencoders (MAE)~\cite{he2022masked} pre-trained on ImageNet1k~\cite{deng2009imagenet} achieving the highest classification accuracy, 65.29\%.

    This remote sensing dataset, FireRisk, provides a new direction for fire risk assessment, and we make it publicly available on \url{https://github.com/CharmonyShen/FireRisk}.
\end{abstract}


\section{Introduction}
Forests have always been a crucial part of ecosystems because of their capacity to filter air, preserve the quality of the soil, and hold onto precipitation~\cite{morancho2003hedonic}. 
Therefore, if a wildfire extensively burns the forest, it will cause irreparable economic losses as well as harm to the biological ecology.
For instance, Australia has been ravaged by bushfires for more than six months, starting in September 2019. Property loss from these fires is estimated to be worth over \$100 billion. More serious is the deterioration of soil and air quality, and the loss of numerous animals as a result of this bushfire~\cite{haque2021wildfire}.

Many studies have been conducted to address the harmful effects of this natural hazard using a multitude of techniques.
Most existing fire risk modelling is founded on geoscientific knowledge. 
A research conducted by~\cite{jaiswal2002forest} found a strong connection between fire risk models based on geospatial data and actual fire incidences.
Based on this theoretical investigation, many traditional fire risk models are developed from fire-related parameters utilizing various data analysis techniques.
For example,~\cite{akay2017gis} generated forest fire risk maps using the multiple-criteria decision analysis approach.

Due to the proliferation of satellite and aerial remote sensing projects that are available to the public, enormous remote sensing images are now more accessible.
Moreover, as a result of the development of optical sensors, the resolution of remote sensing photographs has grown substantially, allowing surface features to be differentiated more clearly.
In past few years, remote sensing images have been widely used in many practical tasks\cite{castelluccio2015land,fassnacht2016review,liu2020local}.
Therefore, several research have examined incorporating remote sensing images into fire risk assessments.
As in an earlier study by~\cite{huesca2009assessment}, seasonal remote sensing data from MODIS satellite imagery, climate data and fuel type were combined to discuss the seasonal fire potential in different regions using the fire potential index (FPI).
Some recent studies utilize machine learning methods for remote sensing imagery and related geographic variables. Using geographical information given by Landsat 8 satellite images such as land surface temperature (LST), normalized differential moisture index (NDMI), and land use and land cover (LULC),~\cite{kalantar2020forest} predicted the vulnerability to forest fires with three basic machine learning models.
In addition,~\cite{quan2021corrigendum} suggested a topography-, weather-, and fuel-based fire assessment approach in which the fuel variables were derived from the MODIS remote sensing project and random forests were used to investigate the association between variables and wildfire in order to build a dataset of wildfire potential.
However, these solutions still rely to some extent on other geoscientific features, which need specialised knowledge.
Although the use of geospatial data can improve the accuracy of assessing fire risk as much as possible, these models lack adequate generality due to inconsistencies in fire risk features between models.
Thus, we question whether a simpler method exists for linking solely remote sensing images to fire risk while still achieving satisfactory results.

\begin{figure}
    \centering
    \includegraphics[width=\linewidth]{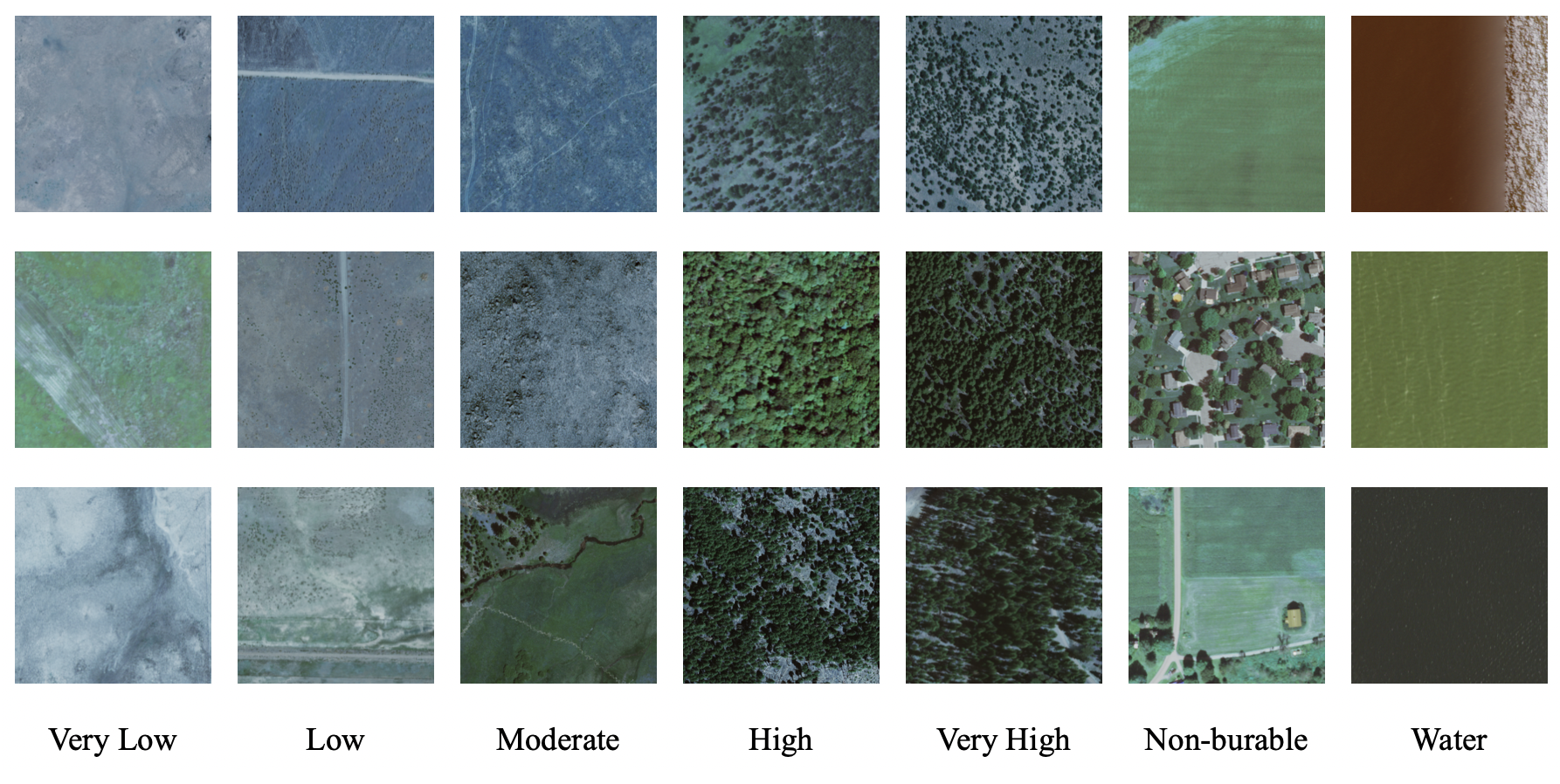}
    \caption{This overview shows sample images of all 7 labels in our FireRisk. The images measure $270\times270$ pixels, with a total of 91,872 images.}
    \label{fig:1}
\end{figure}

Because of these motivations, it is assumed that remote sensing images contain geographic information that can reflect the degree of fire risk, e.g., the species of trees in the forest and the proximity to human activity areas identified from remote sensing images can indirectly mirror the difficulty of fire occurrence.
The objective of this work is to develop a simple scheme to describe a mapping between remote sensing imagery and fire risk on the Earth's surface.
Using data provided by Wildfire Hazard Potential (WHP)~\cite{dillon2020wildfire} and National Agriculture Imagery Program (NAIP)~\cite{maxwell2017land}, we construct a remote sensing dataset, FireRisk, for fire risk assessment.
This dataset consists of 91,872 labelled images, where each remote sensing image corresponds to a fire risk class, with a total of 7 fire risk classes.
Fig.~\ref{fig:1} depicts an example image for each of these classes, from which it is intuitively evident that forest cover and fire danger are correlated.
In addition, we collect the unlabelled dataset from NAIP used to pre-train for the following self-supervised benchmark models.
The labelled fire risk dataset, FireRisk, and the unlabelled dataset for pre-training, UnlabelledNAIP, are publicly available on 
\url{https://github.com/CharmonyShen/FireRisk}.

Further, we provide benchmark evaluations of supervised and self-supervised learning on FireRisk.
Using transfer learning, we fine-tune ResNet-50~\cite{he2016deep}, ViT-B/16~\cite{dosovitskiy2020image}, as well as DINO~\cite{caron2021emerging} and MAE~\cite{he2022masked} with ViT-B/16 as the backbone, all of which were pre-trained on ImageNet~\cite{deng2009imagenet}, using our FireRisk to evaluate the classification accuracy and F1-score of different benchmarks.
For the self-supervised learning models, DINO~\cite{caron2021emerging} and MAE~\cite{he2022masked}, we additionally measure the end-to-end performance of benchmark models pre-trained on our UnlabelledNAIP.

Our main contributions of this work are:
\begin{itemize}
    \item We propose FireRisk, a remote sensing dataset for fire risk assessment, and offer a novel method for constructing a mapping between 7 fire risk classes and remote sensing images.
    \item To investigate the performance of supervised and self-supervised learning on our FireRisk, we employ ResNet~\cite{he2016deep}, ViT~\cite{dosovitskiy2020image}, DINO~\cite{caron2021emerging}, and MAE~\cite{he2022masked} as benchmark models.
    With the use of transfer learning, we obtain the results of these models pre-trained on ImageNet~\cite{deng2009imagenet} and then fine-tuned on our FireRisk.
    \item Using the performance of our benchmarks on 20\% and 50\% of the training data from the original FireRisk, we illustrate the efficiency of data labelling in FireRisk as well as the sensitivity of various benchmark models to the amount of labelled data.
    \item We gather an unlabelled dataset, UnlabelledNAIP, from the NAIP remote sensing project and utilize it to pre-train novel latent representations of DINO~\cite{caron2021emerging} and MAE~\cite{he2022masked}.
    The results of fine-tuning on FireRisk using these two representations demonstrate the potential of different self-supervised benchmarks for enhancement in fire risk assessment.
\end{itemize}

\section{Related Work}

Our work focuses on proposing a remote sensing image classification dataset for fire risk assessment.
In this context, we first review the related remote sensing datasets.
In addition, 4 advanced supervised and self-supervised learning benchmarks are implemented to evaluate our FireRisk, hence some similar computer vision approaches are also presented in this section.

\subsection{Remote Sensing Classification Datasets}

Using satellites or aeroplanes, remote sensing is the technique of detecting and monitoring the physical features of a region by measuring its optical radiation from a distance.
These remote sensing images, collected using special cameras, can help researchers 'sense' what is occurring on the earth~\cite{campbell2011introduction}.
Since remote sensing images contain some implicit geographical features, they are frequently utilized to address practical classification tasks, such as land-use classification~\cite{castelluccio2015land}, climatic zone classification~\cite{liu2020local}, and tree species classification~\cite{fassnacht2016review}.
Thus, several labelled remote sensing datasets, including BigEarthNet~\cite{sumbul2019bigearthnet}, EuroSAT~\cite{helber2019eurosat}, and So2Sat~\cite{zhu2019so2sat}, have emerged to train deep learning models in computer vision to solve various classification tasks of remote sensing images.

The majority of the existing remote sensing image datasets for wildfires focus on identifying fires that have occurred from remote sensing images, for example,~\cite{de2021active} created a remote sensing dataset for fires based on Landsat-8 images and used the Convolutional Neural Network (CNN) based U-Net to detect active fires on the surface.
Regarding fire risk assessment tasks, most of the explored approaches~\cite{huesca2009assessment,kalantar2020forest,quan2021corrigendum} emphasise the combination of remote sensing images and geographic information; there are no publicly available datasets for fire risk classification using only remote sensing images.

\subsection{Computer Vision Models}

Based on whether labelled training data is required, deep learning models in computer vision can be broadly classified into, supervised and unsupervised models, where self-supervised learning is also a sort of unsupervised learning.

\paragraph{Supervised Learning} The key to supervised learning is the extraction of features from image information, which can be divided into two main schools of thought as follows:

\textit{CNNs Architecture.} Convolutional structures have been introduced into computer vision for image classification since they use dimensionality reduction to lower the number of parameters while preserving the relative locations of pixels.
\cite{krizhevsky2017imagenet} first introduced, AlexNet, a deep CNN architecture, which is more effective compared to traditional manual extraction features on ImageNet.
Theoretically, boosting the CNN’s depth can improve its performance. This is because deep networks incorporate features and classifiers of multiple dimensions in a multi-layered end-to-end manner, and the deeper the network structure, the richer the level of features.
Nevertheless, increasing the network's depth may lead to problems such as vanishing gradients, exploding gradients and network degradation.
To address these problems, ResNet was proposed by~\cite{he2016deep} to enable the training of deeper networks by introducing residual blocks.

\textit{Transformers Architecture.} In recent years, as self-attention-based mechanisms have become more prevalent in natural language processing~\cite{vaswani2017attention}, numerous Transformer-based systems have created waves in computer vision because to the efficiency and scalability of Transformers.
Some studies, such as Detection Transformer (DETR)~\cite{carion2020end}, attempt to merge CNN with Transformer, while others have completely replaced the CNN framework with Transformer, such as Vision Transformer (ViT)~\cite{dosovitskiy2020image}, which utilize the encoder structure comprised of multi-headed self-attention blocks in the Transformer to extract features and then employ MLP for image classification.

\paragraph{Self-supervised Learning} As the complexity of deep learning models increases, data hunger has been a hurdle for supervised models to overcome, while self-supervised learning methods can automatically learn latent representations from unlabelled data.
In computer vision, it can be categorized into predictive, generative and contrastive methods, and the latter two of which are explained in detail below:

\textit{Generative Methods.} Generative methods learn latent representations for self-supervised learning by reconstructing or generating input data~\cite{wang2022self}.
\cite{ballard:modular} initially presented the notion of autoencoder, which employs an encoder to map the input to a latent vector and then a decoder to reconstruct the input from the vector.
To improve the robustness to noise in the data,~\cite{vincent2010stacked} suggested a denoising autoencoder.
Inspired by the outstanding performance of ViT~\cite{dosovitskiy2020image} for feature extraction, MAE~\cite{he2022masked} reconstructed randomly masked patches using the structure of denoising autoencoder.

\textit{ Contrastive Methods.} Contrastive methods train the model by comparing inputs that are semantically equivalent, such as two augmented views of the same image.
Yet, over emphasis on the similarity between input pairs may result in model collapse.
The most intuitive approach is to introduce negative samples~\cite{oord2018representation,tian2020contrastive}, while other approaches use teacher-student networks to transfer knowledge in both network structures without negative samples.
\cite{grill2020bootstrap} proposed BYOL, the first method for self-supervised learning based on knowledge distillation.
DINO~\cite{caron2021emerging} further explored the introduction of ViT backbone into knowledge distillation, they built a teacher network with a dynamic encoder and avoided model collapse by centering the output of the teacher network.

Although many advanced supervised and self-supervised deep learning models have advanced performance, they are difficult to train on a relatively small amount of training data due to their large parameter size.
A common solution is to use transfer learning, which applies knowledge acquired from generic or well-studied datasets to new downstream tasks.
In the field of remote sensing images, transfer learning has strong applicability for both supervised and self-supervised models, with its ability to fine-tune the models, pre-trained on general image datasets, for specific remote sensing tasks.

\section{Datasets}

Our work focuses on constructing a remote sensing dataset, FireRisk, for fire risk assessment based on the fire risk levels provided by the WHP project~\cite{dillon2020wildfire}.
In addition, we also supply an unlabelled dataset, UnlabelledNAIP, that contains remote sensing images for pre-training self-supervised benchmarks.

\subsection{Construction of Our FireRisk Dataset}

\paragraph{Extracting Labels From the WHP} The construction of our FireRisk is inspired by the WHP~\cite{dillon2020wildfire}, a raster dataset developed by the U.S. Department of Agriculture that provides a relatively authoritative geographic assessment of fire risk and the intensity of wildfires in the United States.
Their model was developed using a series of geostatistical data, including spatial estimates of wildfire susceptibility and intensity generated by FSim~\cite{finney2011simulation}, spatial fuels and vegetation data from LAND-FIRE~\cite{rollins2009landfire}, and fire occurrence point locations from the FPA~\cite{finney2011simulation}.

Based on their investigation, we utilize their classified 2020 version of the WHP raster dataset~\cite{dillon2020wildfire}, containing 7 fire risk levels, from which we extract fire risk labels for the area represented by each raster.
We download the data from the Missoula Fire Sciences Laboratory's official website \footnote{\url{https://www.firelab.org/project/wildfire-hazard-potential}} in .gdb format, a geodatabase format that divides the United States, including Hawaii, Alaska, and the continental United States, into grids with 270-meter sides and provides information on the fire risk level of the area represented by each grid.
Since the coordinate information of the raster dataset exists implicitly, we use the geoprocessing tools in ArcGIS, a geospatial information processing program, to transform the raster dataset into point features that are exported as tabular data containing only the coordinates of each geographic grid and its corresponding labels.

Because the WHP~\cite{dillon2020wildfire} contains a large number of point features and geographically adjacent areas may have similar fire risk features, only a subset of the WHP with 110 data intervals of equally spaced sampling for all data points is employed in this study.

\paragraph{Construction Our Dataset Using the NAIP} After obtaining the fire risk labels for each grid, sufficient remote sensing imagery is required to construct our FireRisk remote sensing dataset.
With the large number of satellite image projects publicly available on the Google Earth Engine platform\footnote{\url{https://developers.google.com/earth-engine/datasets/catalog/USDA_NAIP_DOQQ}}~\cite{gorelick2017google}, we can easily access remote sensing projects from different time periods.
However, since each grid in the WHP raster dataset covers only a 270-meter-square area, the lower resolution remote sensing images in commonly used satellite remote sensing projects, such as Landsat~\cite{loveland2012landsat}, MODIS~\cite{justice2002modis}, and Sentinel~\cite{malenovsky2012sentinels}, do not provide sufficient geographic information within each grid area.

By the utilization of an aerial platform to capture orthorectified remote sensing images of the Earth's surface, the NAIP project, presented by~\cite{maxwell2017land}, with its use of an aerial platform to acquire orthorectified remote sensing images of the Earth's surface can achieve high spatial resolution, compared with these projects.
Although the images are collected independently by each state and their spatial attributes may vary, all images in the current NAIP project are available at a minimum resolution of 1 meter.
To optimize image quality and restrict shadow length, the sun must be 30 degrees above the horizon during image capture, and the cloud cover cannot exceed 10\% per quarter of the remote sensing image patches.
In addition, because the primary purpose of NAIP~\cite{maxwell2017land} is agricultural mapping, the images are collected during the plant growing season, with no snow or flood coverage allowed.

\begin{figure}
  \centering
  \includegraphics[width=\linewidth]{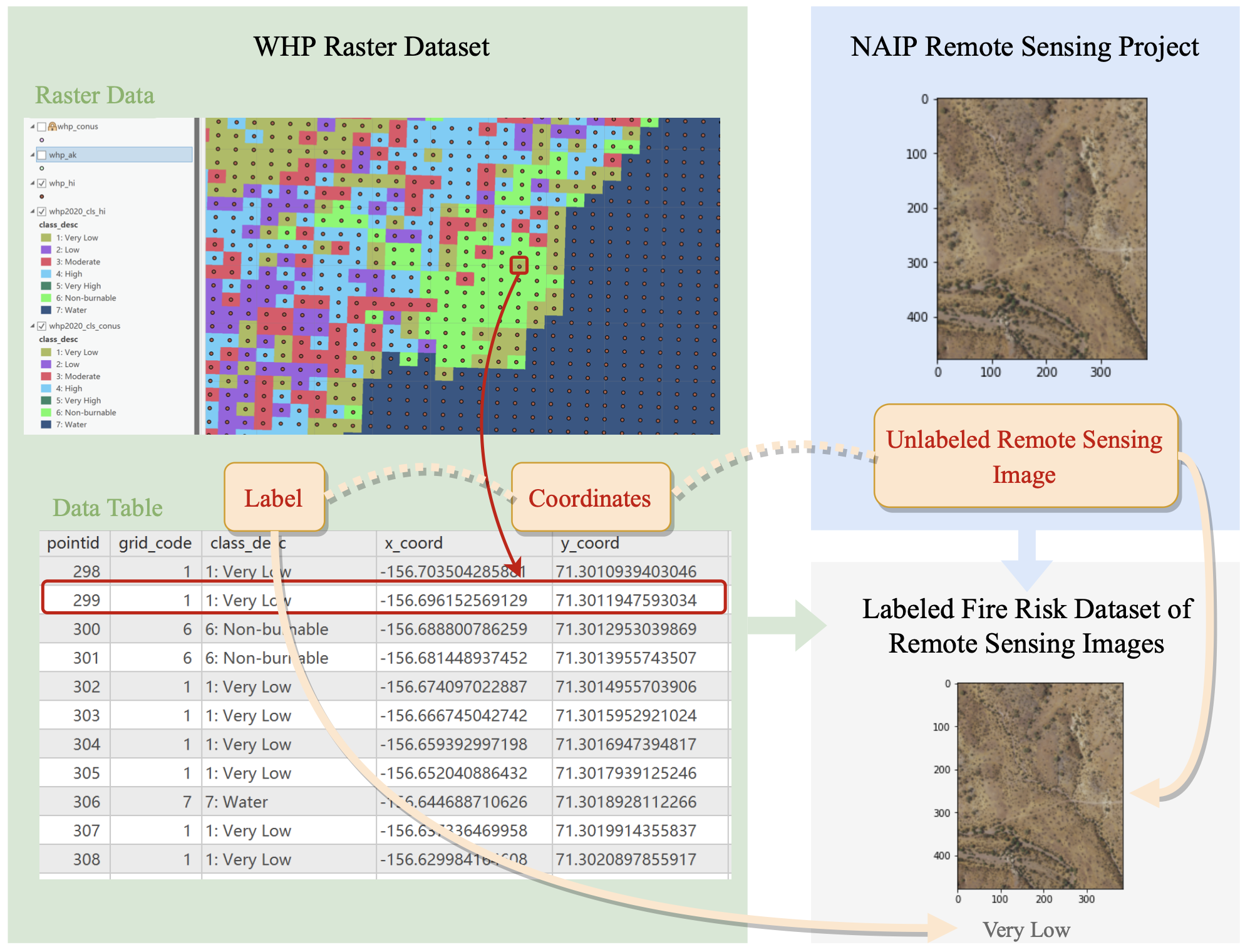}
  \caption{The diagram illustrates the workflow of establishing the FireRisk. To build our FireRisk, the center coordinates of a square area of $270m\times270m$ and the fire risk labels are collected from the WHP raster dataset~\cite{dillon2020wildfire}, followed by obtaining the remote sensing images from the NAIP project~\cite{maxwell2017land} based on the coordinates.}
  \label{fig:2}
\end{figure}

Therefore, our work utilizes NAIP~\cite{maxwell2017land} to collect remote sensing images.
First, the coordinates of the four vertices of the grid with a fire risk label are derived from the center coordinates of each grid in the WHP dataset~\cite{dillon2020wildfire} using some basic geographic coordinate transformation.



Subsequently, we access the remote sensing images in the NAIP~\cite{maxwell2017land} dataset based on the grid coordinates using Google Earth Engine~\cite{gorelick2017google}.
In the spectral configuration, only the R, G, and B channels of each remote sensing image are extracted, with each channel represented by an 8-bit unsigned integer.
In the temporal configuration, since the NAIP project~\cite{maxwell2017land} inherently has a relatively long revisit period, we set the time span for image sampling from January 1, 2019, to December 1, 2020, in order to obtain valid remote sensing images.
In theory, the remote sensing images obtained from the NAIP project~\cite{maxwell2017land} should be square. However, in practice, the majority of remote sensing images are near-square rectangles due to slight angles between the aircraft sensors and the ground, and inconsistent longitude and latitude resolutions in some states.
So finally, we crop the center of each remote sensing image to a square image of $270\times270$ pixels.

Our work, as seen in Fig.~\ref{fig:2}, follows this workflow to construct a remote sensing dataset for a fire risk assessment.
Thus, the fire risk label of a grid derived from the WHP raster dataset~\cite{dillon2020wildfire} can be linked by geographic coordinates to the remote sensing image in the NAIP remote sensing project~\cite{maxwell2017land}, resulting in a mapping connection between the remote sensing image and its corresponding fire risk label.

\begin{figure}
  \centering
  \includegraphics[width=\linewidth]{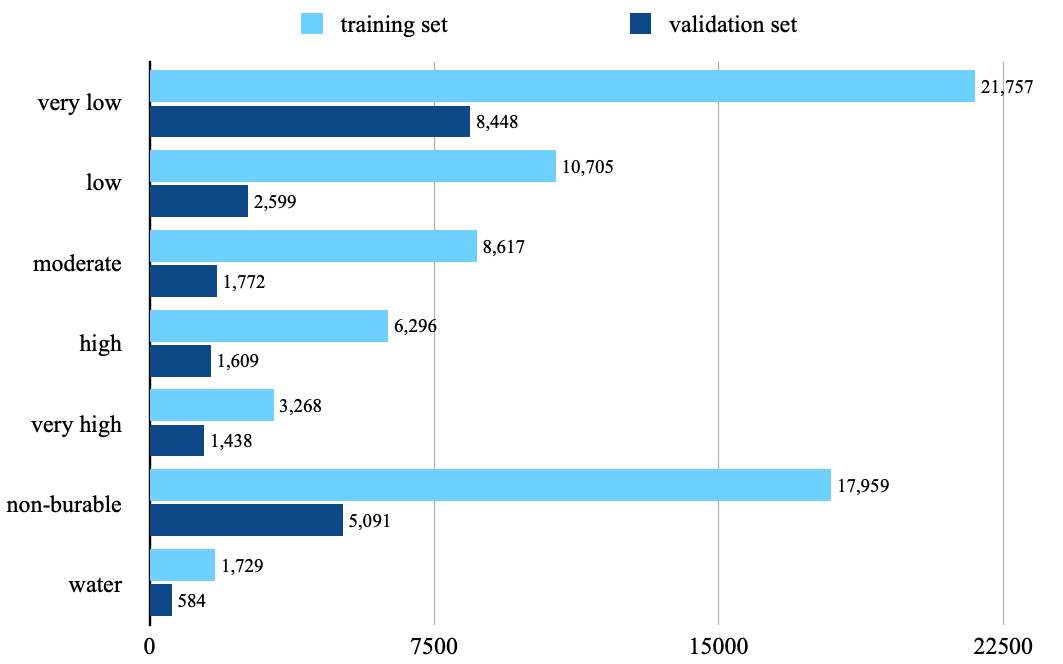}
  \caption{Data distribution of the training and validation sets of our FireRisk on the labels of 7 fire risk levels.}
  \label{fig:3}
\end{figure}

\begin{figure*}
  \centering
  \includegraphics[width=0.86\linewidth]{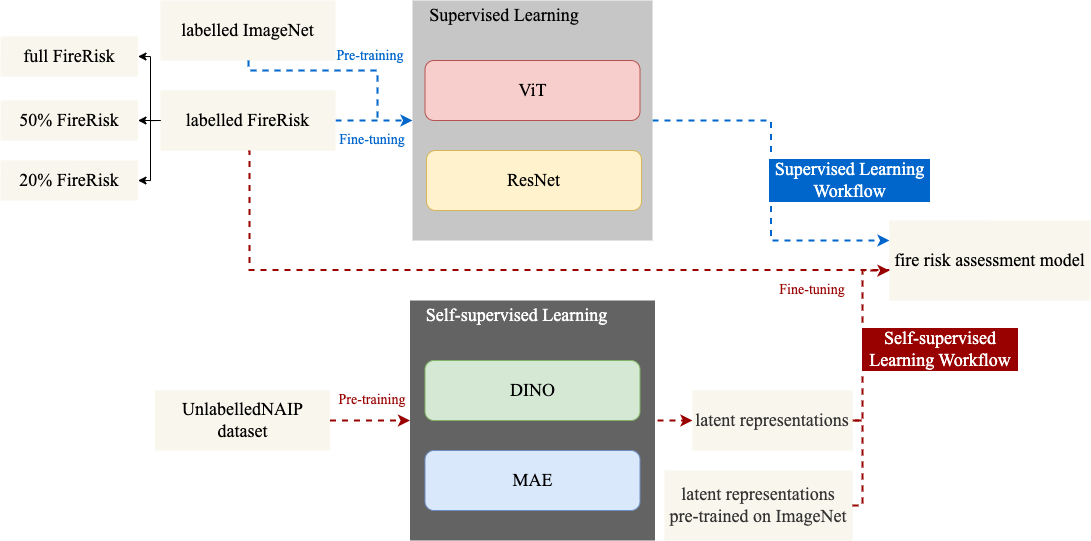}
  \caption{Workflow of our supervised and self-supervised benchmarks. The supervised process represented by the blue arrow trains the models on our FireRisk using pre-trained weights. The self-supervised process denoted by the red arrow is split into two alternatives: fine-tuning the models on our FireRisk using existing latent representations or using pre-trained latent representations on our UnlabelledNAIP.}
  \label{fig:4}
\end{figure*}

As illustrated in Fig.~\ref{fig:3}, our FireRisk contains 7 classes according to the fire risk levels. There are a total of 91,872 labelled remote sensing images, with 70,331 serving as the training set and 21,541 as the validation set.

\subsection{Our UnlabelledNAIP for Pre-training}

In addition, in order to improve the applicability of our self-supervised learning benchmark to our FireRisk, we gather 199,976 NAIP~\cite{maxwell2017land} remote sensing images from Google Earth Engine~\cite{gorelick2017google} for pre-training to generate latent representations.
This unlabelled dataset, UnlabelledNAIP, has the same image size and time period filtering as our FireRisk to allow the self-supervised models to learn as many features as possible from the same style of remote sensing imagery.

\section{Benchmarks}

To evaluate the benchmark performance of our FireRisk for the fire risk assessment task, we validate it in two dimensions: supervised and self-supervised learning.
Fig.~\ref{fig:4} describes the overall workflow for assessing fire risk in our work, where for the supervised learning benchmark we use the methods of ResNet~\cite{he2016deep} and ViT~\cite{dosovitskiy2020image}, while for the self-supervised learning, we select two representative self-supervised models for their performance, namely DINO~\cite{caron2021emerging} and MAE~\cite{he2022masked}.

\subsection{Supervised Learning}

Our dataset provides a remote sensing image classification task labelled with fire risk levels, which, like other classification tasks in computer vision, can usually be predicted with labelled data using supervised learning.
Digital images often consist of a vast number of pixels, and each pixel comprises several channels, making it difficult to generalize the original features when dealing with a multitude of parameters in an image task.

One option for extracting image features is to employ a convolutional structure.
Considering that ResNet~\cite{he2016deep} is widely used in deep learning due to its effectiveness in mitigating network degradation, we choose it as one of our supervised benchmarks.

In addition, other studies introduce Transformer structures to overcome the CNN’s lack of holistic grasp of the input data itself due to inductive bias, making it easier to extract long-distance spatial dependencies between global data, the most famous of which, ViT~\cite{dosovitskiy2020image}, represents another direction of development in computer vision.
Thus, we use it as our second supervised benchmark.

Due to the limited size of our FireRisk training set, it is difficult to support the training of a large number of model parameters.
Influenced by the concept of transfer learning, we transfer the parameters trained on large real-world image datasets to our remote sensing dataset.
As shown in the supervised learning workflow in Fig.~\ref{fig:4}, we fine-tune the ResNet~\cite{he2016deep} and ViT~\cite{dosovitskiy2020image}, pre-trained on ImageNet~\cite{deng2009imagenet}, on our FireRisk, respectively, to produce a supervised benchmark performance for fire risk assessment.

\subsection{Self-supervised Learning}

In contrast, self-supervised learning, as a form of representation learning, can learn visual features from a large number of unlabelled images without the involvement of labelled data.

To determine the benchmark performance of the self-supervised learning on our dataset, we investigate the performance of two representative self-supervised models based on the ViT architecture as the backbone, the contrastive learning DINO~\cite{caron2021emerging} based on knowledge distillation and the generative model MAE~\cite{he2022masked} based on autoencoder, on our FireRisk.

For the self-supervised learning process in Fig.~\ref{fig:4}, we adopt two processing schemes to produce the self-supervised benchmark performance, which are:

\textit{Pre-trained on ImageNet~\cite{deng2009imagenet}:} using the latent representation obtained by pre-training on ImageNet~\cite{deng2009imagenet}, and then fine-tuning it on the labelled FireRisk, by which the generalized knowledge in the latent representations can be transferred to the remote sensing imagery domain;

\textit{Pre-trained on Our UnlabelledNAIP:} first pre-training on our UnlabelledNAIP, the unlabelled remote sensing dataset we collect, to obtain the latent representation based on remote sensing imagery, and then fine-tuning it on our labelled FireRisk.

Compared to the former, the latter processing scheme has an additional step of constructing our unique latent representations, that will reflect features of remote sensing imagery more similar to FireRisk.
In generating the latent representations, we use ViT-B/16 as the backbone architecture for MAE~\cite{he2022masked}, pre-trained for 80 epochs on our UnlabelledNAIP, while we train DINO~\cite{caron2021emerging} for 100 epochs because of the slow convergence of DINO on our pre-trained dataset.

\section{Experiments and Evaluation}

\begin{table}[htbp]
\caption{Performance of several supervised and self-supervised models on FireRisk with different sizes. In our implementation, our supervised benchmark uses ResNet-50 and ViT-B/16 and the self-supervised models, DINO and MAE use ViT-B/16 as backbone.}
\begin{center}
\begin{tabular}{|c|c|c|c|c|}
\hline
\textbf{Dataset} & \textbf{Model}& \textbf{pre-trained} & \textbf{Acc.} & \textbf{F1.} \\
\hline
FireRisk & ResNet~\cite{he2016deep} & ImageNet1k~\cite{deng2009imagenet} & 63.20 & 52.56 \\
 & ViT~\cite{dosovitskiy2020image} & ImageNet1k~\cite{deng2009imagenet} & 63.31 & 52.18 \\
 & DINO~\cite{caron2021emerging} & ImageNet1k~\cite{deng2009imagenet} & 63.36 & 52.60 \\
 & DINO~\cite{caron2021emerging} & UnlabelledNAIP & 63.44 & 52.37 \\
~ & \cellcolor{gray!40}MAE~\cite{he2022masked} & \cellcolor{gray!40}ImageNet1k~\cite{deng2009imagenet} & \cellcolor{gray!40}\textbf{65.29} & \cellcolor{gray!40}\textbf{55.49} \\
 & MAE~\cite{he2022masked} & UnlabelledNAIP & 63.54 & 52.04 \\

\hline
50\% & ResNet~\cite{he2016deep} & ImageNet1k~\cite{deng2009imagenet} & 62.09 & 50.27 \\
FireRisk & ViT~\cite{dosovitskiy2020image} & ImageNet1k~\cite{deng2009imagenet} & 62.22 & 50.07 \\
 & DINO~\cite{caron2021emerging} & ImageNet1k~\cite{deng2009imagenet} & 61.75 & 51.21 \\
 & DINO~\cite{caron2021emerging} & UnlabelledNAIP & 62.49 & 51.35 \\
 & \cellcolor{gray!40}MAE~\cite{he2022masked} & \cellcolor{gray!40}ImageNet1k~\cite{deng2009imagenet} & \cellcolor{gray!40}63.70 & \cellcolor{gray!40}50.23 \\
 & MAE~\cite{he2022masked} & UnlabelledNAIP & 62.68 & 52.05 \\

\hline
20\% & ResNet~\cite{he2016deep} & ImageNet1k~\cite{deng2009imagenet} & 61.37 & 49.53 \\
FireRisk & ViT~\cite{dosovitskiy2020image} & ImageNet1k~\cite{deng2009imagenet} & 61.43 & 48.80 \\ 
 & DINO~\cite{caron2021emerging} & ImageNet1k~\cite{deng2009imagenet} & 60.95 & 50.72 \\
 & DINO~\cite{caron2021emerging} & UnlabelledNAIP & 61.96 & 50.83 \\ 
 & \cellcolor{gray!40}MAE~\cite{he2022masked} & \cellcolor{gray!40}ImageNet1k~\cite{deng2009imagenet} & \cellcolor{gray!40}62.51 & \cellcolor{gray!40}51.13 \\
 & MAE~\cite{he2022masked} & UnlabelledNAIP & 61.80 & 50.07 \\
\hline
\end{tabular}
\label{tab:2}
\end{center}
\end{table}

We apply the pre-training weights of different supervised and self-supervised models on our constructed FireRisk and its subsets, and we examine the performance differences between these various benchmark models in two main ways:
one is to validate the efficiency of labelling in FireRisk by evaluating their robustness to datasets of varying sizes by comparing their performance on different subsets;
the other is to compare the impact of the latent representations generated on different pre-trained datasets on the self-supervised models for fire risk assessment.
Table \ref{tab:2} shows the results of this series of experiments.

\textit{Evaluation Metrics:}
As in most multiclass classification tasks, we mainly use accuracy and macro F1-score as evaluation metrics.
While macro F1-score is influenced by a smaller number of classes, it is employed as a complement to accuracy so that the contributions of \textit{High} and \textit{Very High}, which are smaller in number but more significant in reality, are not ignored.

\textit{Experimental Configurations:}
In our implementation, for the supervised models, we use ViT-B/16~\cite{dosovitskiy2020image} and ResNet-50~\cite{he2016deep} pre-trained on ImageNet1k~\cite{deng2009imagenet} respectively to fine-tune the models on FireRisk.
For the self-supervised architectures, MAE~\cite{he2022masked} and DINO~\cite{caron2021emerging}, we use ViT-B/16~\cite{dosovitskiy2020image} as the backbone and fine-tune on FireRisk using latent representations pre-trained on ImageNet1k~\cite{deng2009imagenet} and our UnlabelledNAIP, respectively.
We use 2 GPUs and a batch size per GPU of 16 for training, and take the results of the highest accuracy out of 100 epochs.

\subsection{Overall Analysis}

\begin{figure}
  \centering
  \includegraphics[width=\linewidth]{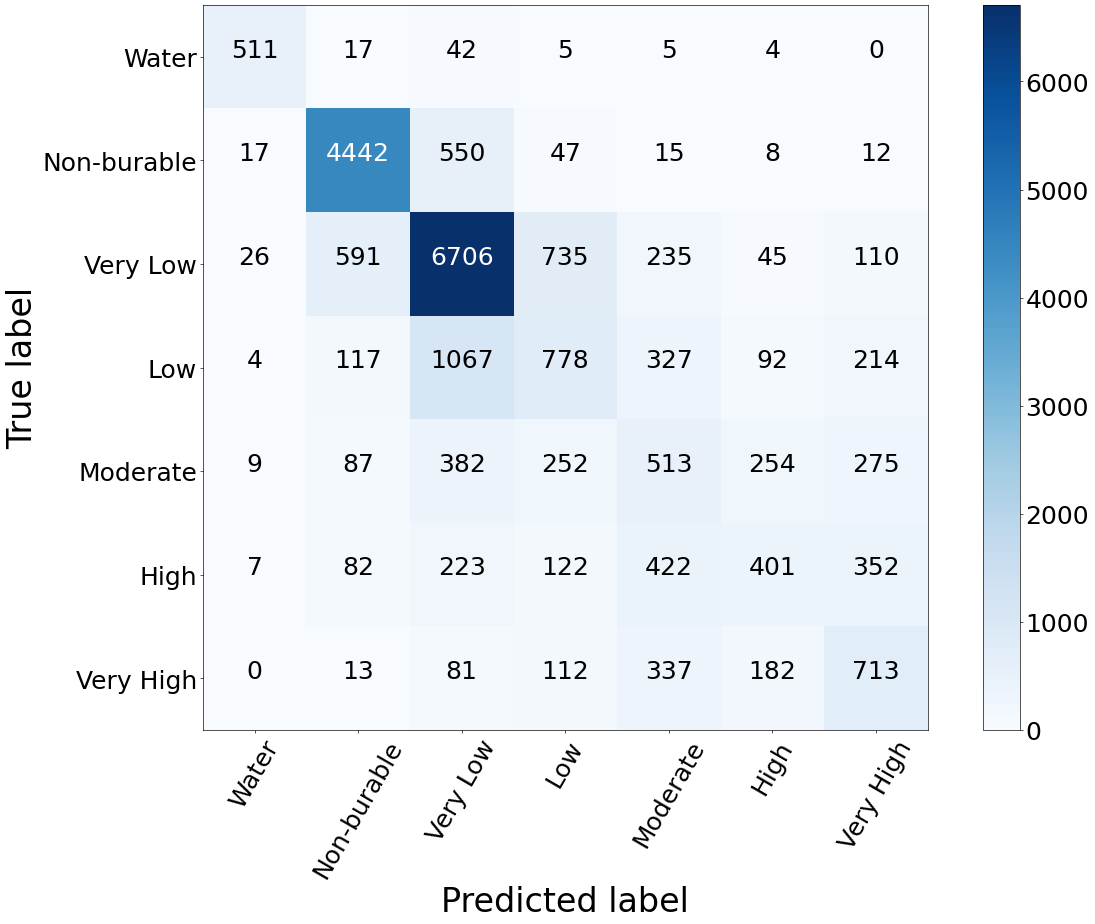}
  \caption{The confusion matrix of the optimal benchmark model, the MAE pre-trained on ImageNet. Due to the ordinal nature of the labels, many misclassifications are seen in cells adjacent to the main diagonal.}
  \label{fig:5}
\end{figure}

\begin{figure*}
  \centering
  \begin{subfigure}{0.45\linewidth}
    \centering
    \includegraphics[width=\linewidth]{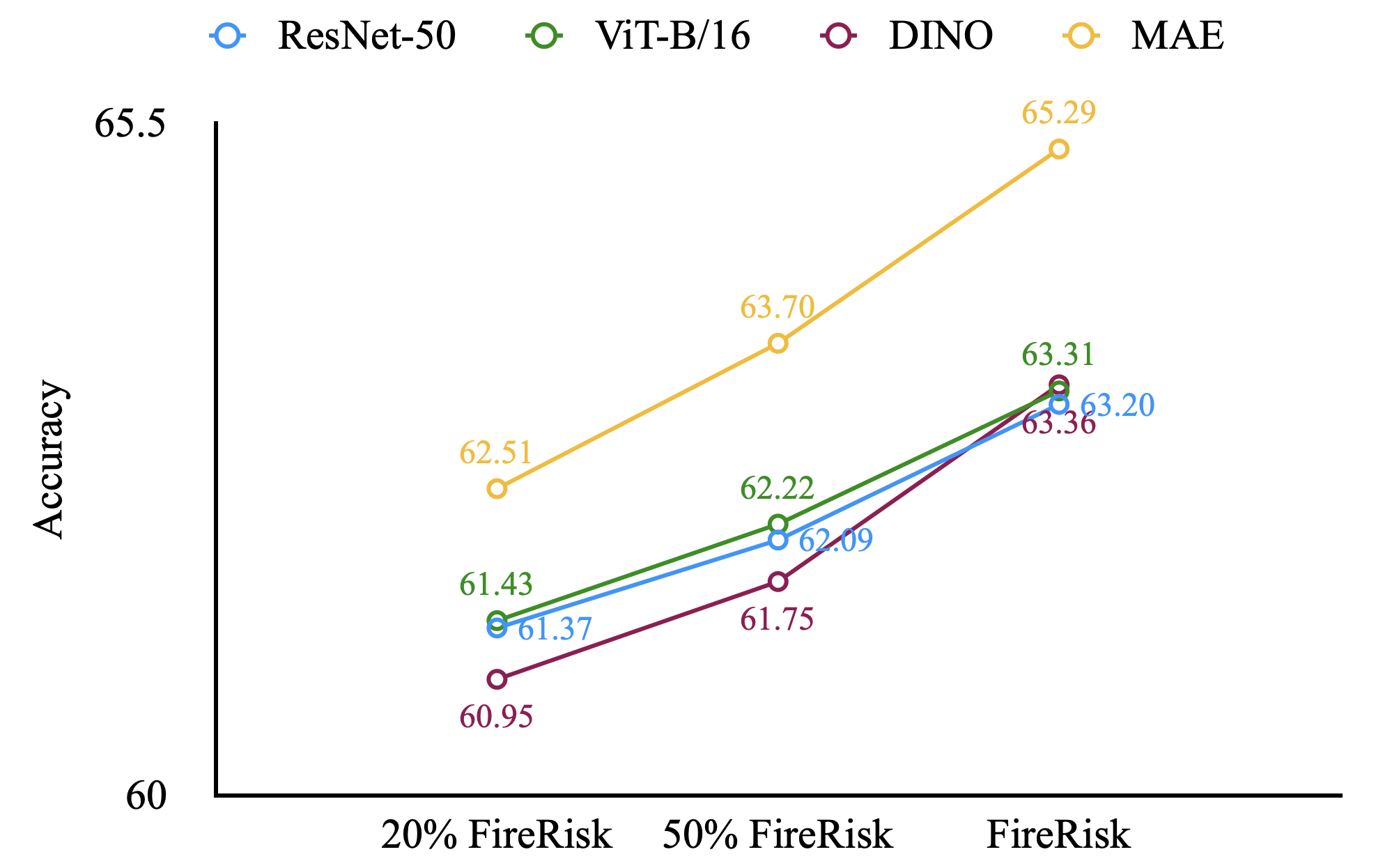}
    \caption{Performance of models on different sizes of FireRisk.}
    \label{fig:6a}
  \end{subfigure}\hspace{10mm}
  \begin{subfigure}{0.45\linewidth}
    \centering
    \includegraphics[width=\linewidth]{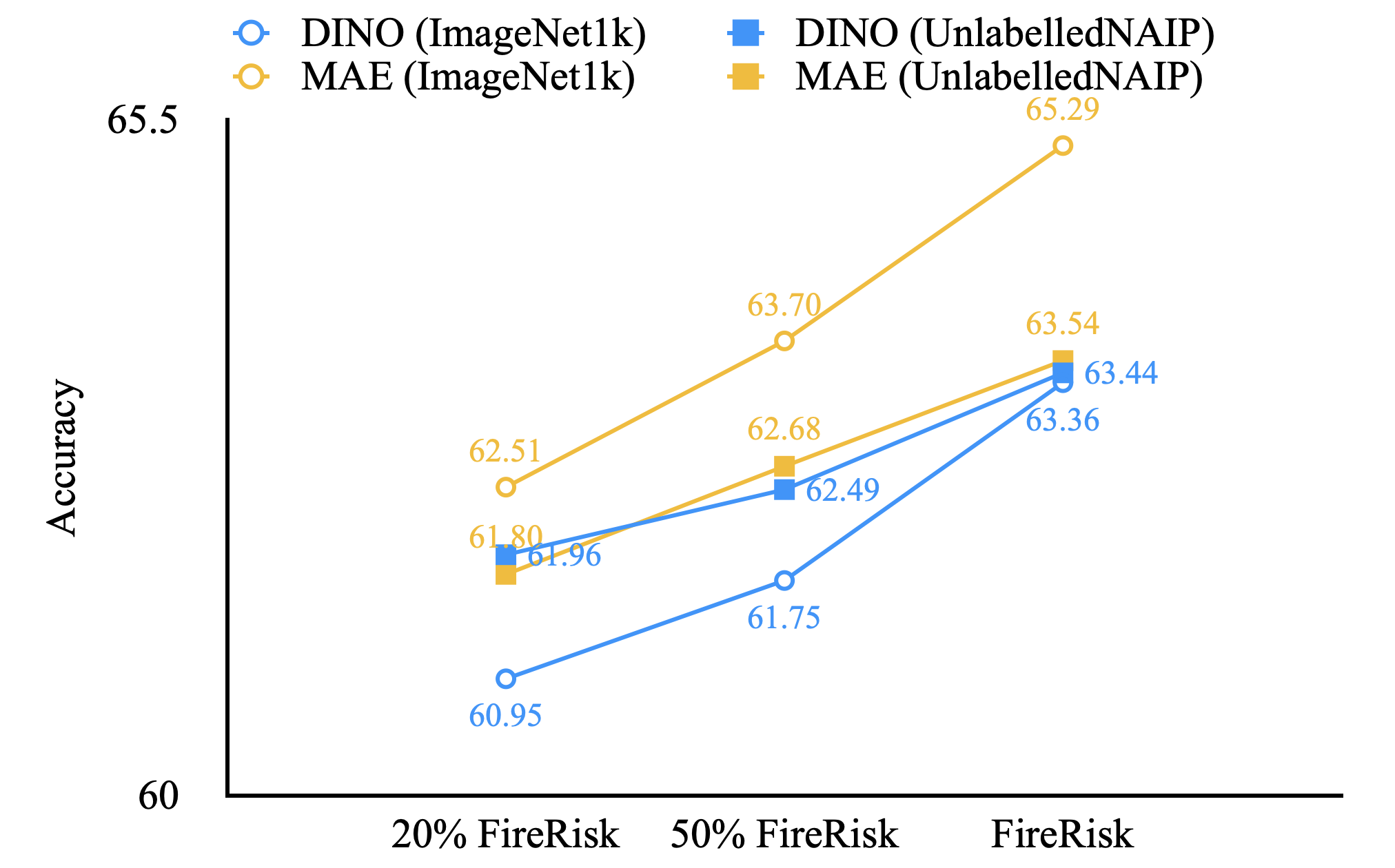}
    \caption{Accuracy differences between different latent representations of the self-supervised benchmark.}
    \label{fig:6b}
  \end{subfigure}
  \caption{Accuracy of different benchmarks and different self-supervised latent representations on different sizes of FireRisk.}
  \label{fig:6}
\end{figure*}

Table \ref{tab:2} shows that the performance of the self-supervised benchmark on FireRisk outperforms the supervised benchmark in general, while the performance of MAE~\cite{he2022masked} is significantly higher than the other models.
This is because the performance of supervised models is overly dependent on labelled information, while in the image domain, images contain much richer internal information than labels provide.
For fire risk assessment tasks, the self-supervised learning approach is better at extracting implicit features in remote sensing images.

Our optimal baseline model is the MAE~\cite{he2022masked} pre-trained on ImageNet~\cite{deng2009imagenet}, whose confusion matrix is shown in Fig.~\ref{fig:5}.
The confusion matrices of the remaining benchmarks have similar features.
Taking the confusion matrix of MAE as an example, we can see that, for the classification accuracy of each class, the highest is \textit{Water} which can reach 87.50\%, while \textit{Non-burnable} and \textit{Very Low} also have high accuracy, but \textit{Low}, \textit{Moderate} and \textit{High} have lower accuracy.
For the label \textit{Low}, where most of the misclassifications are found on the two fire risk classes \textit{Low} and \textit{Very Low}, which can demonstrate that our FireRisk is prone to misleading on these two labels.
The same problem of similar feature ambiguity exists for \textit{Moderate} and \textit{High}.

In practical fire risk assessment tasks, one usually pays more attention to the recall of high fire risk because one needs to screen out the high-risk areas in remote sensing images as accurately as possible in order to prevent wildfires.
For example, in Fig.~\ref{fig:5}, for all remote sensing images labelled \textit{Very High} in the validation set, the MAE~\cite{he2022masked} predicts only 713 images correctly, which accounts for 49.58\%.
However, considering that the images labelled \textit{High} and \textit{Very High} usually have some similar features, people also focus on the area with the fire risk level of \textit{High} in practice.
If the misclassification of \textit{Very High} as \textit{High} is included, the 'recall' of \textit{Very High} can reach 62.24\%.
Through this analysis, our FireRisk can reflect the actual role of FireRisk to some extent.

\subsection{Label Efficiency Evaluation}

In this experiment, in order to investigate label efficiency, we focus on the robustness of these benchmark models on FireRisk obtained with this processing method of ours when the amount of data is reduced.
We sample the training set at 50\% and 20\% from the full training set to obtain two subsets, 50\% FireRisk and 20\% FireRisk, respectively.
For the supervised and self-supervised benchmarks we investigate, we apply the same model configurations to fine-tune the models on this series of datasets and then evaluate on the same validation set.

Integrating all the models pre-trained on ImageNet1k~\cite{deng2009imagenet} in Table \ref{tab:2} according to the results of the same model on different size datasets, we can obtain Fig.~\ref{fig:6a}.
It can be seen that all benchmarks increase as the size of the dataset increases.
The two self-supervised models, DINO~\cite{caron2021emerging} and MAE~\cite{he2022masked}, have a higher increasing trend and are more sensitive to the size of the training data than the supervised benchmarks.
This indicates that more training is required to fit the latent representations generated by self-supervised learning to the features of fire risk in remote sensing images.

\subsection{Fine-tuning of Self-supervised Representations}

For the self-supervised benchmark, we compare the differences in latent representations of these two benchmark models, DINO~\cite{caron2021emerging} and MAE~\cite{he2022masked}, for feature extraction of remote sensing images in FireRisk.
Compared to ImageNet~\cite{deng2009imagenet}, UnlabelledNAIP is smaller in size but has more similar features of remote sensing images to those on FireRisk.
As shown in Fig.~\ref{fig:6b}, for DINO~\cite{caron2021emerging}, the model pre-trained on UnlabelledNAIP is better than that pre-trained on ImageNet~\cite{deng2009imagenet}, while for MAE~\cite{he2022masked}, the result is the opposite.
Thus, DINO~\cite{caron2021emerging} has higher pre-training efficiency to achieve better performance on our FireRisk, while MAE~\cite{he2022masked} usually requires more data pre-training to achieve its optimal results.

\section{Conclusion}

In this paper, to demonstrate the feasibility of using only remote sensing images to capture fire risk features, we develop a novel dataset, FireRisk, to assess fire risk.
To construct this remote sensing dataset, we obtain the label of fire risk level and their corresponding geographic coordinates for a 270-meter-square area from the WHP~\cite{dillon2020wildfire} raster dataset, and then gather the remote sensing image at this geographic location using the NAIP remote sensing project.

To investigate the benchmark performance of our FireRisk for supervised and self-supervised learning, we employ the advanced CNN-based ResNet~\cite{he2016deep} and attention-mechanism-based ViT~\cite{dosovitskiy2020image} as our supervised benchmarks, and DINO~\cite{caron2021emerging} and MAE~\cite{he2022masked} as our self-supervised benchmarks, respectively.
We also explore the potential of the self-supervised benchmark model by generating new latent representations on UnlabelledNAIP, an unlabelled image dataset we gather.
Furthermore, we validate the efficiency of the labels by analyzing the differences in the robustness of our benchmarks to variations in the amount of training data, using sub-datasets generated by randomly sampling 20\% and 50\% from the training set.

On our FireRisk, the maximum accuracy for the supervised benchmarks can reach 63.31\%, while for the self-supervised benchmarks, the MAE~\cite{he2022masked} pre-trained on ImageNet1k~\cite{deng2009imagenet} can achieve the optimal accuracy of all models at 65.29\%.
It is demonstrated that our self-supervised learning benchmarks outperform supervised learning on FireRisk, although their improvement on less training data is limited.
Our new pre-trained latent representations are also complementary to the self-supervised representations, and our representation of DINO~\cite{caron2021emerging} has a considerable increase, which can reach 63.44\% compared to 63.36\% for the DINO pre-trained on ImageNet~\cite{deng2009imagenet}.

The FireRisk dataset proposed in this work confirms the validity of using only remote sensing data for fire assessment, and has a simpler implementation process and better generalization than traditional fire assessment approaches.

\section*{Acknowledgment}
\small
\vspace{-0.35em}
This research was supported by The University of Melbourne’s Research Computing Services and the Petascale Campus Initiative. 
This research was supported (partially or fully) by the Australian Government through the Australian Research Council's Centre of Excellence for Children and Families over the Life Course (Project ID CE200100025).

\vspace{-1.7em}
{\small
\bibliographystyle{ieee_fullname}
\bibliography{egbib}
}

\end{document}